\documentclass[runningheads]{llncs}

\usepackage{caption}
\usepackage[T1]{fontenc}
\usepackage{tabularx} 
\usepackage{rotating}
\usepackage{makecell}
\usepackage{subcaption}
\usepackage[export]{adjustbox}

\usepackage{graphicx}

\begin{document}

\title{Design of a Modular Mobile Inspection and Maintenance Robot for an Orbital Servicing Hub}

\titlerunning{Modular Mobile Inspection and Maintenance Robot}

\author{Tianyuan Wang\inst{1}\orcidID{0000-0003-3545-4937} \and
Mark A Post\inst{1}\orcidID{0000-0002-1925-7039} \and
Mathieu Deremetz\inst{2}}

\authorrunning{T. Wang, M. Post, M. Deremetz}

\institute{University of York, Heslington, York, England, YO10 5DD
\email{\{tianyuan.wang,mark.post\}@york.ac.uk}\\
\url{https://www.york.ac.uk/physics-engineering-technology/} \and
Space Applications Services NV/SA, Leuvensesteenweg 325, 1932 Sint-Stevens-Woluwe, Belgium\\
\email{mathieu.deremetz@spaceapplications.com}
\url{https://www.spaceapplications.com/}}

\maketitle              
\begin{abstract}
The use of autonomous robots in space is an essential part of the ``New Space'' commercial ecosystem of assembly and re-use of space hardware components in Earth orbit and beyond.  The STARFAB project aims to create a ground demonstration of an orbital automated warehouse as a hub for sustainable commercial operations and servicing.  A critical part of this fully-autonomous robotic facility will be the capability to monitor, inspect, and assess the condition of both the components stored in the warehouse, and the STARFAB facility itself.  This paper introduces ongoing work on the STARFAB Mobile Inspection Module (MIM).  The MIM uses Standard Interconnects (SI) so that it can be carried by Walking Manipulators (WM) as an independently-mobile robot, and multiple MIMs can be stored and retrieved as needed for operations on STARFAB.  The MIM carries high-resolution cameras, a 3D profilometer, and a thermal imaging sensor, with the capability to add other modular sensors.  A grasping tool and torque wrench are stored within the modular body for use by an attached WM for maintenance operations.  Implementation and testing is still ongoing at the time of writing.  This paper details the concept of operations for the MIM as an on-orbit autonomous inspection and maintenance system, the mechanical and electronic design of the MIM, and the sensors package used for non-destructive testing.
\end{abstract}
\vspace{-18pt}
\section{Introduction}
\vspace{-6pt}
Outer space has become easier to access than ever before with the advent of low-cost commercial launch technologies and improvements in space technology, communications and robotics.  However, the current practice of simply launching single-use bespoke missions from Earth is not economical and sustainable, and the development of In-space Service, Assembly and Maintenance (ISAM) technologies has taken priority to facilitate this new ecosystem.  The ``New Space'' or ``Big Space'' model that is now evolving envisions the use of modular space technologies with universal interoperability that can be assembled, deployed, retrieved, and maintained fully or partly autonomously through the use of robotics in Earth orbit and beyond.  With this model comes the need for in-space robotic storage and handling of required items and resources, which we propose can be fulfilled in the form of an ``Orbital Hub'' that serves as a unified warehousing, assembly, and maintenance facility to which servicer spacecraft could dock and launch carrying with them satellites composed of modular components that can be serviced and reconfigured.  To this end, the project ``A Space Warehouse Concept and Ecosystem to Energize European ISAM (STARFAB)'' \cite{horizon-starfab} introduces a novel concept, the Orbital Automated Warehouse Unit (also known as the Orbital Depot), which serves as a backbone and enabler for sustainable ISAM commercial activities \cite{deremetz2024starfab}.  The introduction of capability to warehouse, reconfigure, and redeploy modular spacecraft components on-orbit brings with it the necessity to regularly inspect and maintain components remotely and at least semi-autonomously. The Inspection and Maintenance (I\&M) capabilities within STARFAB focus on the detection, diagnosis, and reporting of malfunction or damage to facilities, systems, and contents. 
Non-destructive testing (NDT) methods provide the lifecycle analysis with failure prediction and diagnosis capability needed to plan the deployment and operation of modular spacecraft.  STARFAB also needs the capability to carry out light maintenance interventions robotically and on demand, to implement planned or unplanned maintenance of the facilities.  Complex interventions are not considered at present due to the high fundamental complexity of robotic maintenance activities on-orbit.  The Inspection and Maintenance robot for STARFAB is designed as a self-contained spacecraft module called the Mobile Inspection Module (MIM). The MIM uses HOTDOCK (HD) Standard Interconnects (SI) so that it can be carried by Walking Manipulators (WM) and other mobile subsystems \cite{letier2024hotdock}.  It also operates like any other self-contained modular element in the system with independent power and processing resources, and multiple MIMs can be stored and retrieved as needed for operations on STARFAB.
\vspace{-12pt}
\section{System Design}
\vspace{-6pt}
\subsubsection{Requirements}
Table \ref{tab:IMRequirement} summarizes the Inspection and Maintenance requirements for STARFAB.  The main inspection and maintenance challenges for STARFAB are due to requiring autonomous operation in a minimally-controlled vacuum and irradiated environment with extremes of temperature. This limits feasible NDT techniques at present to remotely-deployable optical sensing methods applicable to inspecting a wide range of materials and parts.  Based on this set of requirements, the MIM is designed to carry a high-resolution machine vision camera with encircling illumination unit, optical 3D scanner, and thermal imaging camera.  Maintenance activities are accomplished by using an additional WM to attach and manipulate a grasping tool and torque wrench.
\begin{table}[tb]
\caption{Inspection and Maintenance Requirements}
\label{tab:IMRequirement}
    \begin{tabularx}{\textwidth}{|p{2cm}|X|}
    \hline
    \textbf{Element} &  \textbf{Requirement}\\
    \hline
    Payloads & Automated NDT is possible for all payloads on all external surfaces. \\
    \hline
    Spacecraft & Hosted spacecraft shall be inspected externally without disassembly. \\
    \hline
    Profilometry & Scratch/deformation damage detected from debris larger than $0.3mm$. \\
    \hline
    Resolution & Debris impact damage detectable down to $0.6mm$ diametric size. \\
    \hline
    Reliability & Damage features detected with $90\%$ probability to confidence of $95\%$. \\
    \hline
    Range & Inspection within a range of $0.2m$ to $2m$ from the sensor array. \\
    \hline
    Thermal & Inspection distinguishes localized temperatures of $-40^\circ C$ to $150^\circ C$. \\
    \hline
    Illumination & Ring-topology visible spectrum illumination source is carried for vision. \\
    \hline
    Handling & Replacement parts handled and placed using dedicated robotic tools. \\
    \hline
    Grasping & Gripper tool can grasp objects with dimensions from $0.5cm$ to $10cm$. \\
    \hline
    Torque & Torque tool compatible to NASA PGT tool one; $2.7Nm$ to $30Nm$. \\
    \hline
    \end{tabularx}
\end{table}
\vspace{-16pt}
\subsubsection{Concept of Operations}
The primary use case for the MIM is the inspection of modular spacecraft components known as Orbital Replacement Units (ORUs) and components while they are on route to or from the primary structure. The MIM must be capable of being positioned at any stage of the ORU’s movement, while aiming to keep the primary structure unobstructed.  The secondary use case for the MIM is inspection of the structural integrity of STARFAB itself, from both interior and exterior positions, and that of docked servicer spacecraft.  The MIM can be be positioned by ``walking'' over designated fixture points on the structure with WMs, transported by attaching via a WM to a shuttle system, or attached in place to a fixture point via a WM as an ``eye in hand'' sensor.  
\vspace{-12pt}
\begin{figure}[htb]
    \centering
    \includegraphics[width=0.8\linewidth]{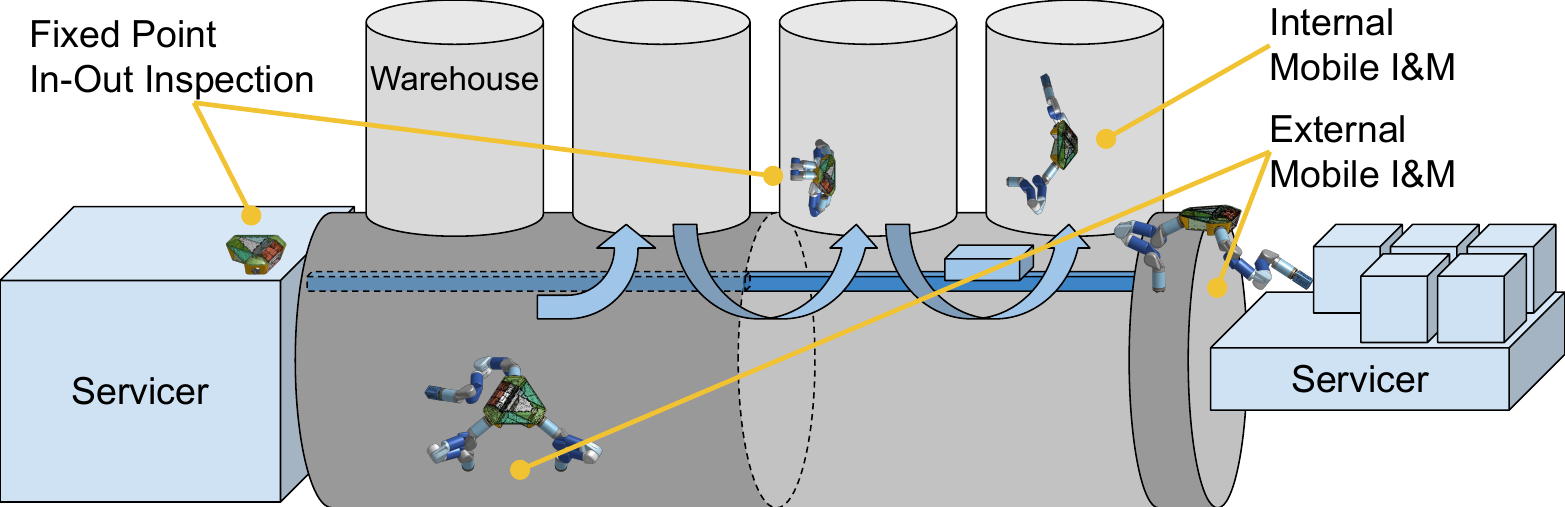}
    \caption{Operations of the MIM performing I\&M inside and outside STARFAB.}
    \label{fig:Operation_Example}
\end{figure}
\vspace{-38pt}
\section{Mechanical Design}
\vspace{-10pt}
\subsubsection{Form Factor}
\begin{figure}[tb]
    \centering
    \begin{subfigure}[b]{0.45\textwidth}
        \centering
        \adjincludegraphics[width=\textwidth,trim={0 {.03\height} 0 {.05\height}},clip]{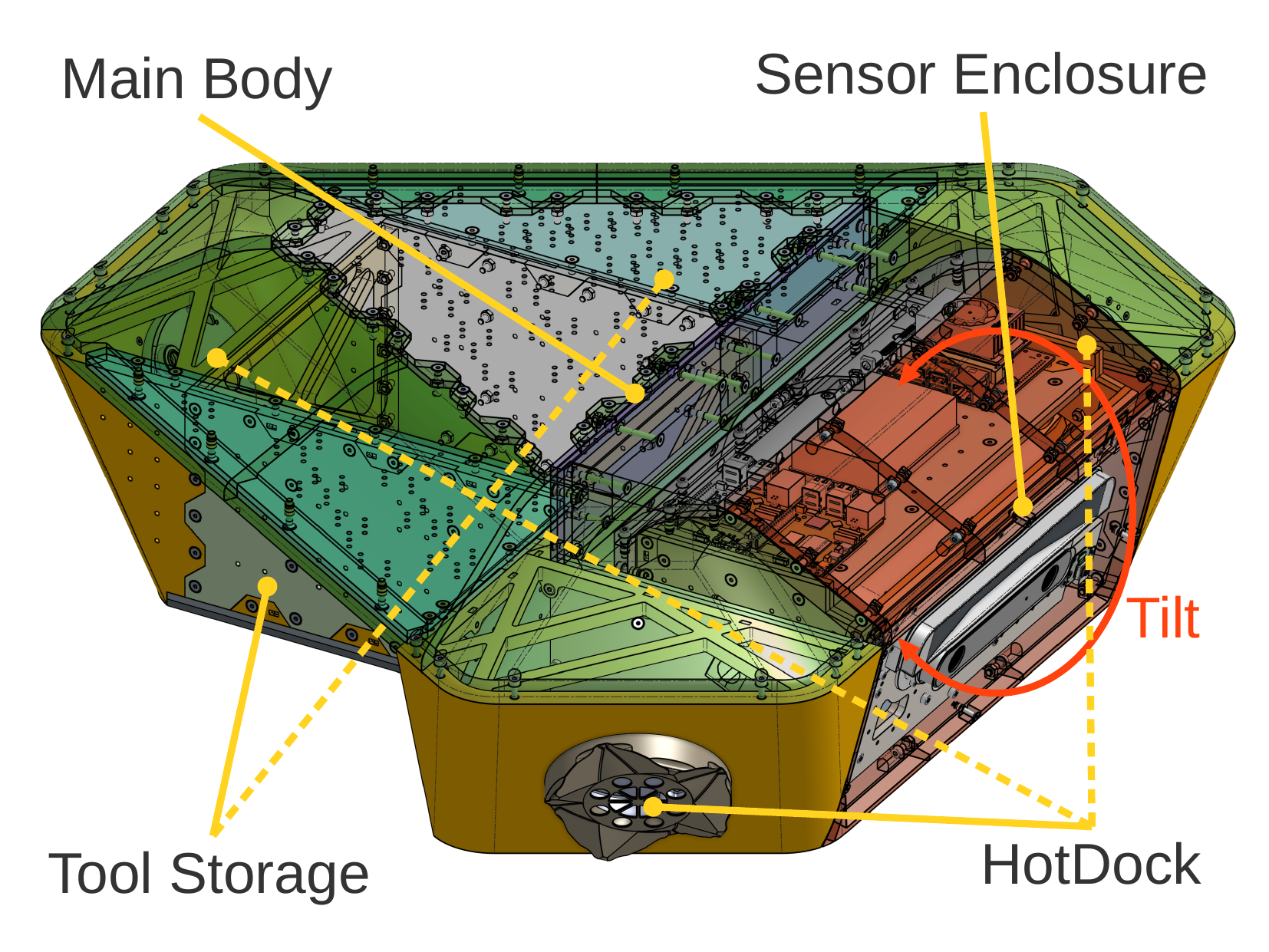}
        \caption{}
        \label{fig:MIM_only_labelled}
    \end{subfigure}
    \begin{subfigure}[b]{0.45\textwidth}
        \centering
        \adjincludegraphics[width=\textwidth,trim={0 {.03\height} 0 {.05\height}},clip]{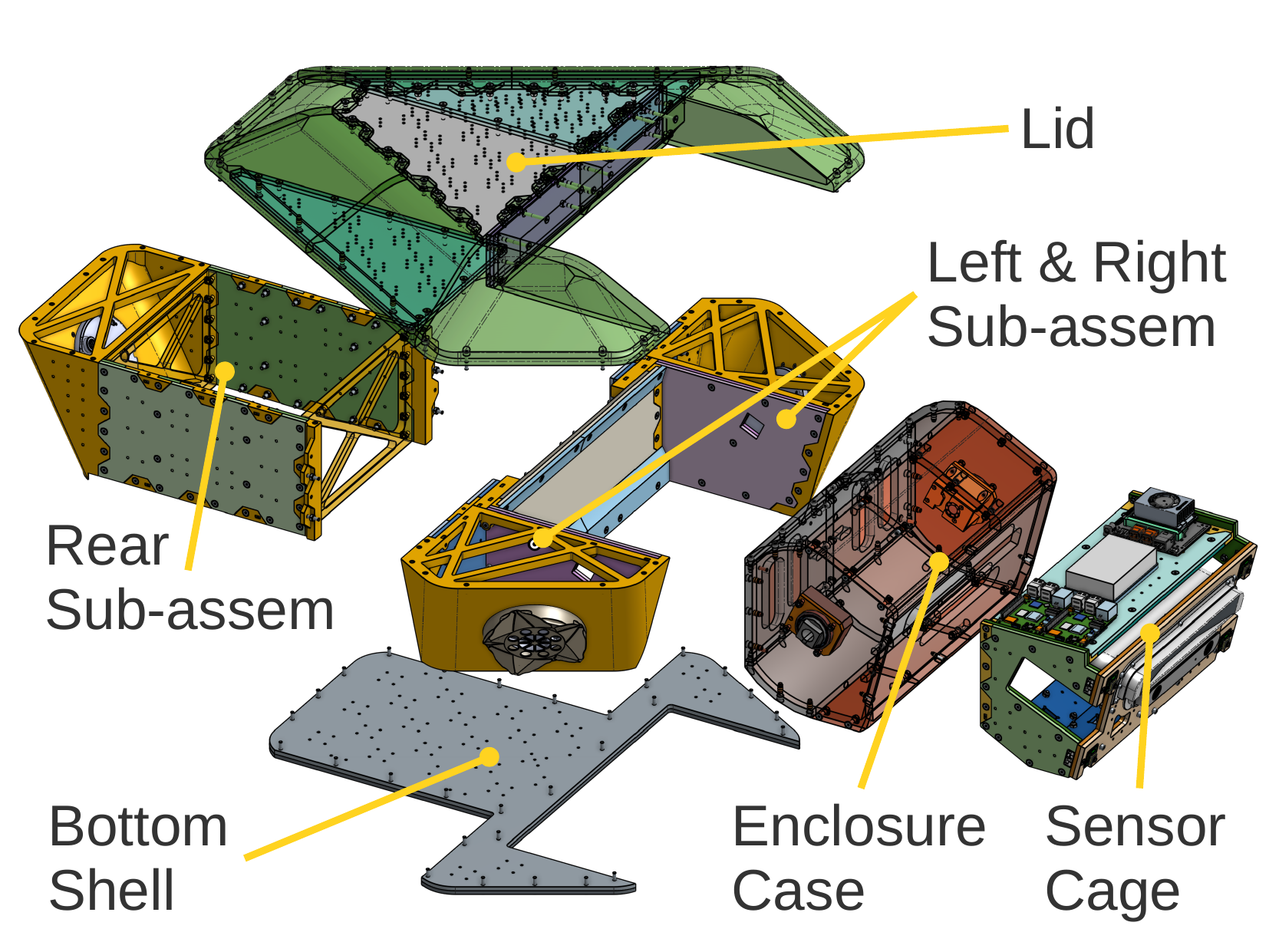}
        \caption{}
        \label{fig:MIM_exploded_labelled}
    \end{subfigure}
    \caption{Mechanical design of the MIM. (\textbf{a}) An overview of the MIM with HDs attached. (\textbf{d}) An exploded view of the main constituent parts of the MIM.}
    \label{fig:MIM_mechanical_design}
\end{figure}
The form factor of the MIM is based on the Multi-Arms Robot (MAR) developed for on-orbit large telescope assembly \cite{deremetz2022design}.  Design underwent three major iterations with careful consideration given to the I\&M requirements, as well as ensuring compatibility with the STARFAB’s existing system, which eventually led to 
a final iteration that aligned with the MAR form factor, shown in Figure \ref{fig:MIM_mechanical_design}, 
to achieve several objectives. First, 
adopting the MAR torso's form factor allows the MIM to utilize the same kinematic configuration, ensuring seamless integration with the existing system. Second, the evenly distributed three HDs, when paired with the WM, provide enhanced flexibility, where the third HD can be positioned as needed to interface with tools.
The MAR torso's top view features a hexagonal shape with unequal sides. The MIM’s sensor components are mounted on one of the longer sides within a cylindrical-like enclosure capable of tilt movement. When the sensor interface is oriented forward, two WMs are mounted on the left and rear HDs, allowing for autonomous movement, 
while the HD on the right side is reserved for an additional robotic arm, either a WM or a lighter alternative, for tool retrieval, manipulation, and storage. This configuration enables the MIM to interact with tools through an independent robotic arm, providing greater degrees of freedom. 
The tools are stored within the open-sided compartments on the two sides of the MIM main body under the lid. These compartments take space both internally and externally to the original MAR torso profile to achieve a balance between the WM-to-tool interference and the tool carrying capacity.
\vspace{-18pt}
\subsubsection{Interfacing}
\begin{figure}[tb]
    \centering
    \begin{subfigure}[b]{0.2\textwidth}
        \centering
        \includegraphics[width=\textwidth]{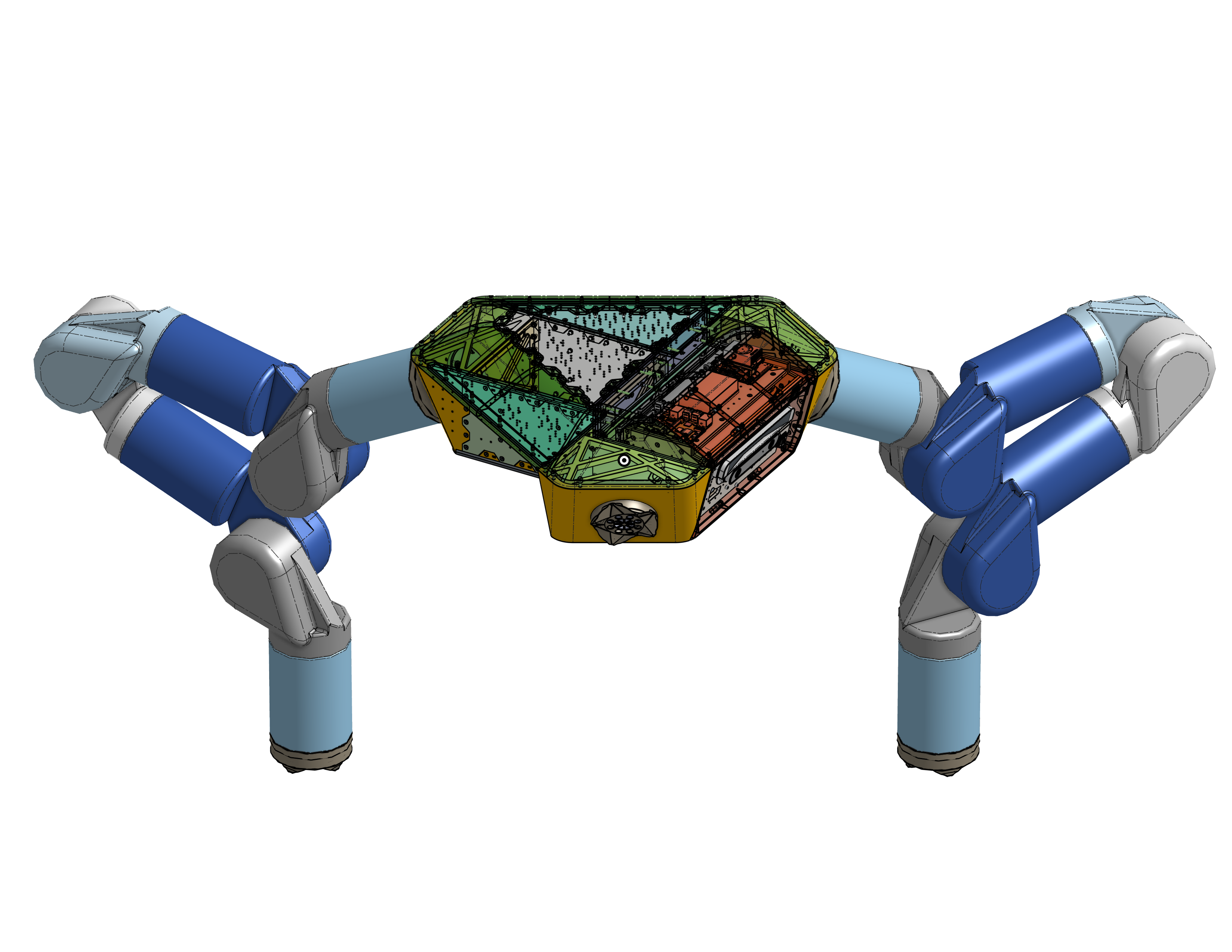}
        \caption{}
        \label{fig:MIM_2_WM_deployed}
    \end{subfigure}
    \begin{subfigure}[b]{0.2\textwidth}
        \centering
        \includegraphics[width=\textwidth]{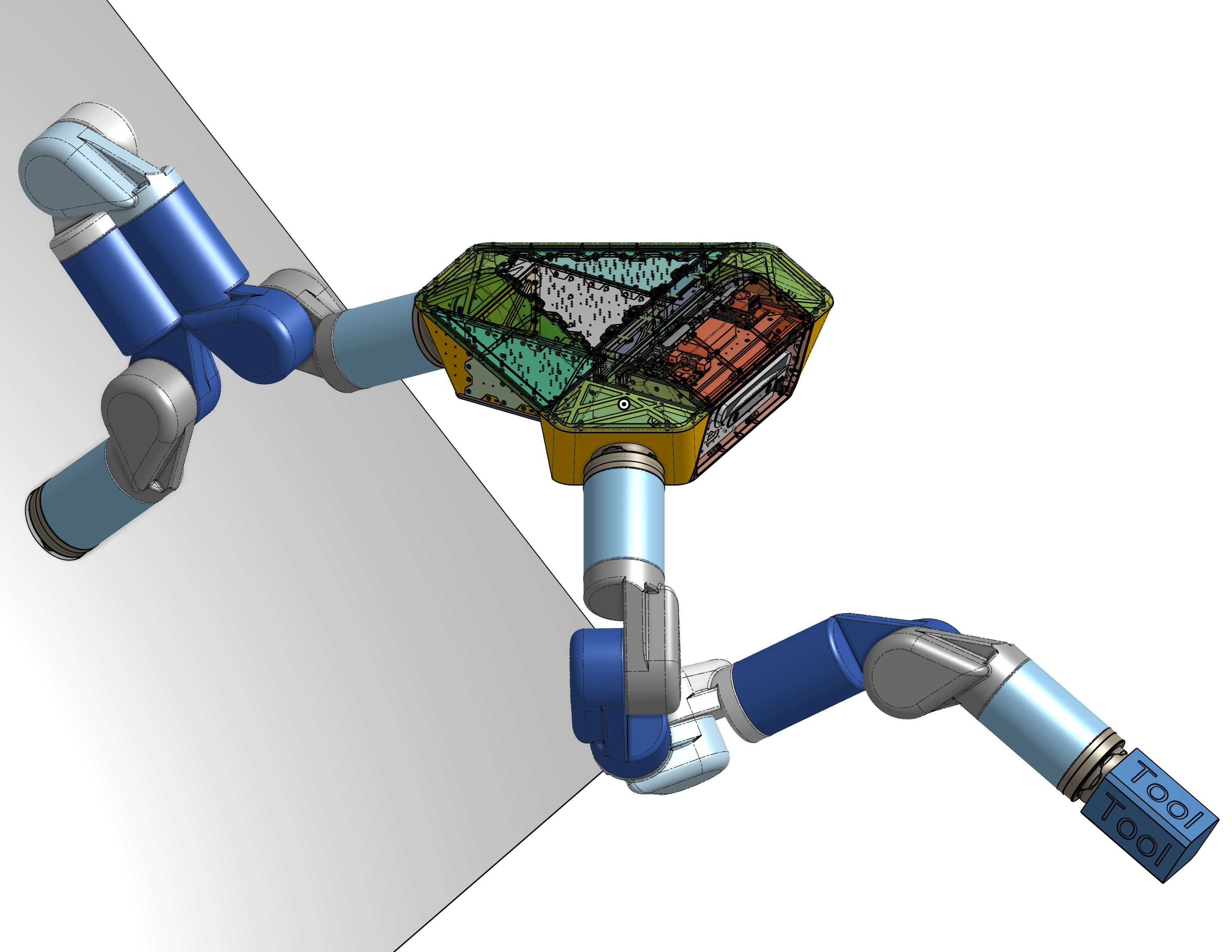}
        \caption{}
        \label{fig:MIM_2_WM_Wall}
    \end{subfigure}
    \begin{subfigure}[b]{0.2\textwidth}
        \centering
        \includegraphics[width=\textwidth]{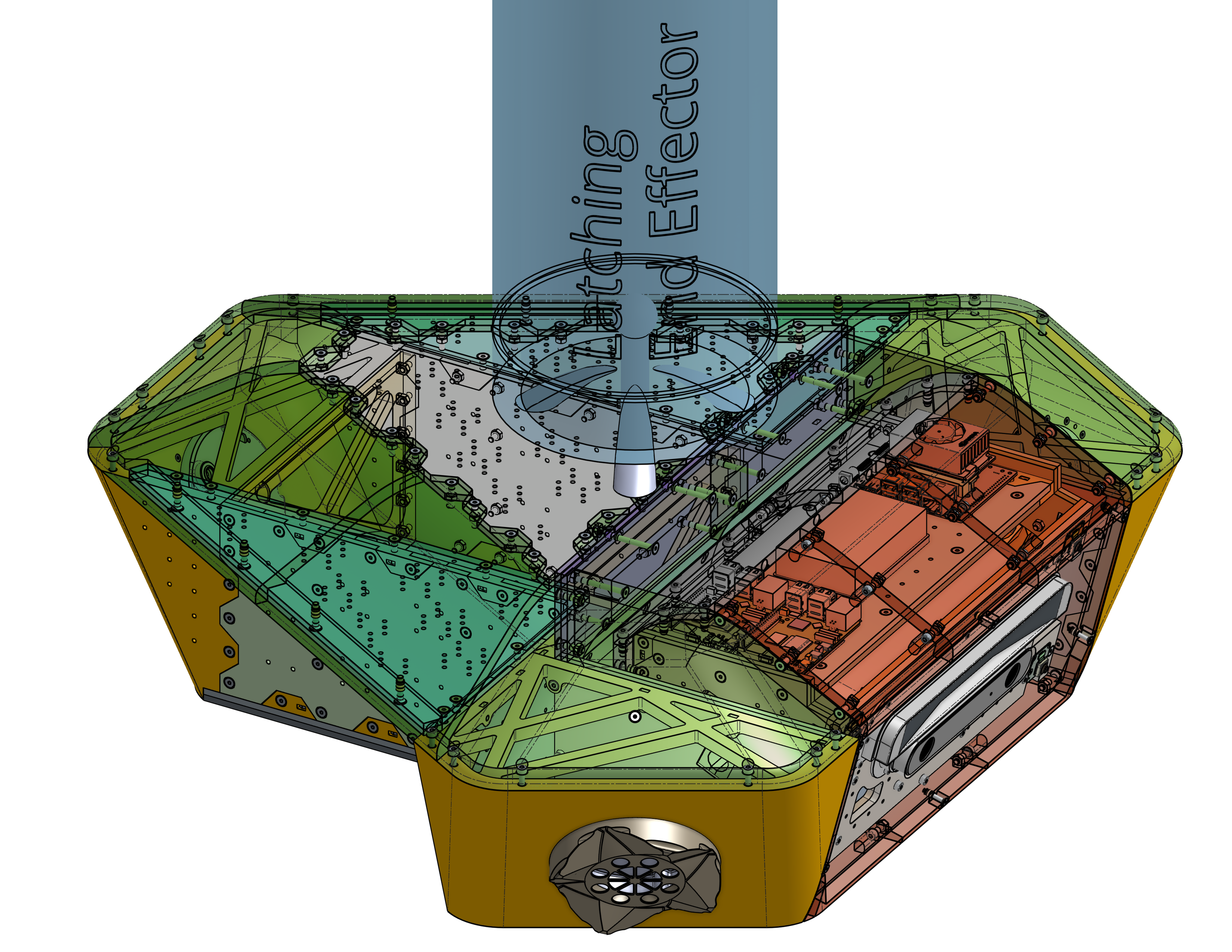}
        \caption{}
        \label{fig:MIM_grapple}
    \end{subfigure}
    \begin{subfigure}[b]{0.2\textwidth}
        \centering
        \includegraphics[width=\textwidth]{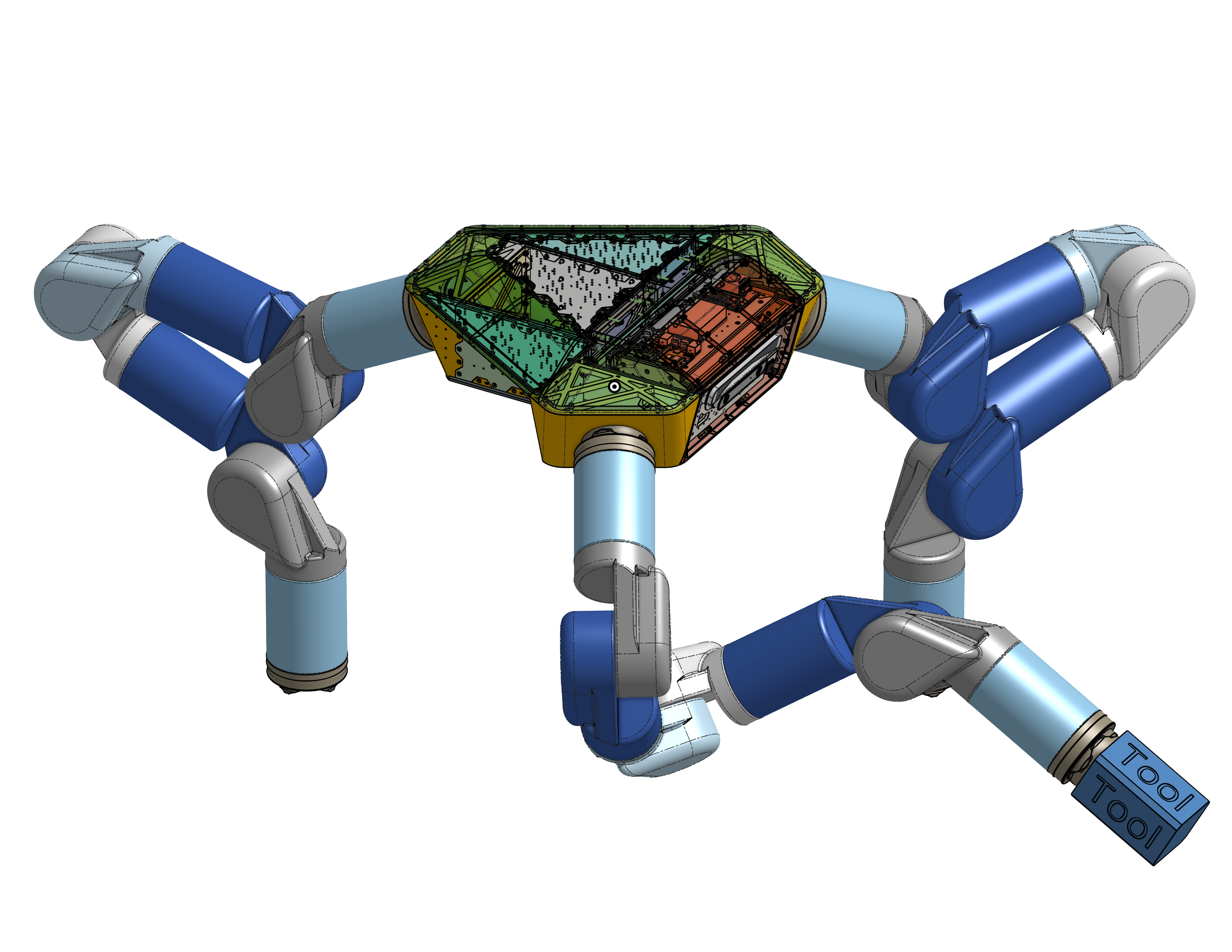}
        \caption{}
        \label{fig:MIM_3_WM_Tool}
    \end{subfigure}
\vspace{-8pt}
    \caption{Example configurations of the MIM. From (\textbf{a}) to (\textbf{d}): walking configuration, externally mounted configuration, large arm mounted configuration, and maintenance MIM Configuration.}
    \label{fig:MIM_configurations}
\end{figure}

The three HDs at the corners of the MIM chassis interface to WMs, payloads, and other modular components. Specifications including quantity and placements of the HDs depend on the configuration of how the MIM is to be integrated with other STARFAB modular systems. As a modular system, the MIM is designed to be flexible enough to fulfil many different roles, as shown in Figure \ref{fig:MIM_configurations}. These include, for example, when operating as an independent robot, it will have two WMs attached, which operate as ``legs'', to walk across infrastructure's attachment points. This allows it to reach external locations on the warehouse unit and areas not easily within reach of manipulators.
When operating as a payload for the WM, with the WM mounted on a shuttle or fixed attachment point, the MIM can be flexibly positioned for inspection using one WM.
When operating as a payload for the external Large Arm of STARFAB, the MIM is held by the grapple fixture and positioned above the area to be inspected.
The manipulation of tools is accomplished by attaching a WM to an unused HD on the MIM 
to access tools in the tool storage. 
Tools are then used by the WM with a camera integrated within the WM operating as an ``eye-in-hand'' sensor and within the view of the MIM’s sensor package, which then operates as an ``eye-to-hand'' sensor.
%
\vspace{-16pt}
\subsubsection{Power}
%
Given the constraints on mass and energy consumption inherent in space applications, the internal electronics of the MIM are designed to be minimalistic, operating at 24V with a power consumption of approximately 5W per unit, except for the on-board computer (OBC) and illumination source. 
Mass reduction remains a primary concern, as the flight version of the MIM is larger and heavier than ground applications. The power bus on the robotic arm operates at 48V, allowing for more efficient power distribution with a current limitation of around 14A for the entire robot.
\vspace{-12pt}
\section{Sensor and Processing Hardware Selection}
\vspace{-12pt}
\subsubsection{On-Board Computing}
The computing hardware for the MIM must have the capacity to run image and signal processing algorithms in real time for I\&M purposes. While future flight hardware and software will be significantly different, for terrestrial testing an OBC running Linux is used to facilitate experimentation with algorithms from the OpenCV machine vision library, Point Cloud Library (PCL), and the InFuse Space Robotics Common Data Fusion Framework (CDFF) \cite{dominguez2020common}.  The central OBC on the MIM is an Nvidia Jetson Orin Nano developer board, which has sensor hardware acceleration in Linux and uses ROS2 over Ethernet as a system bus to other components.
\begin{figure}[tb]
    \centering
    \includegraphics[width=0.8\linewidth]{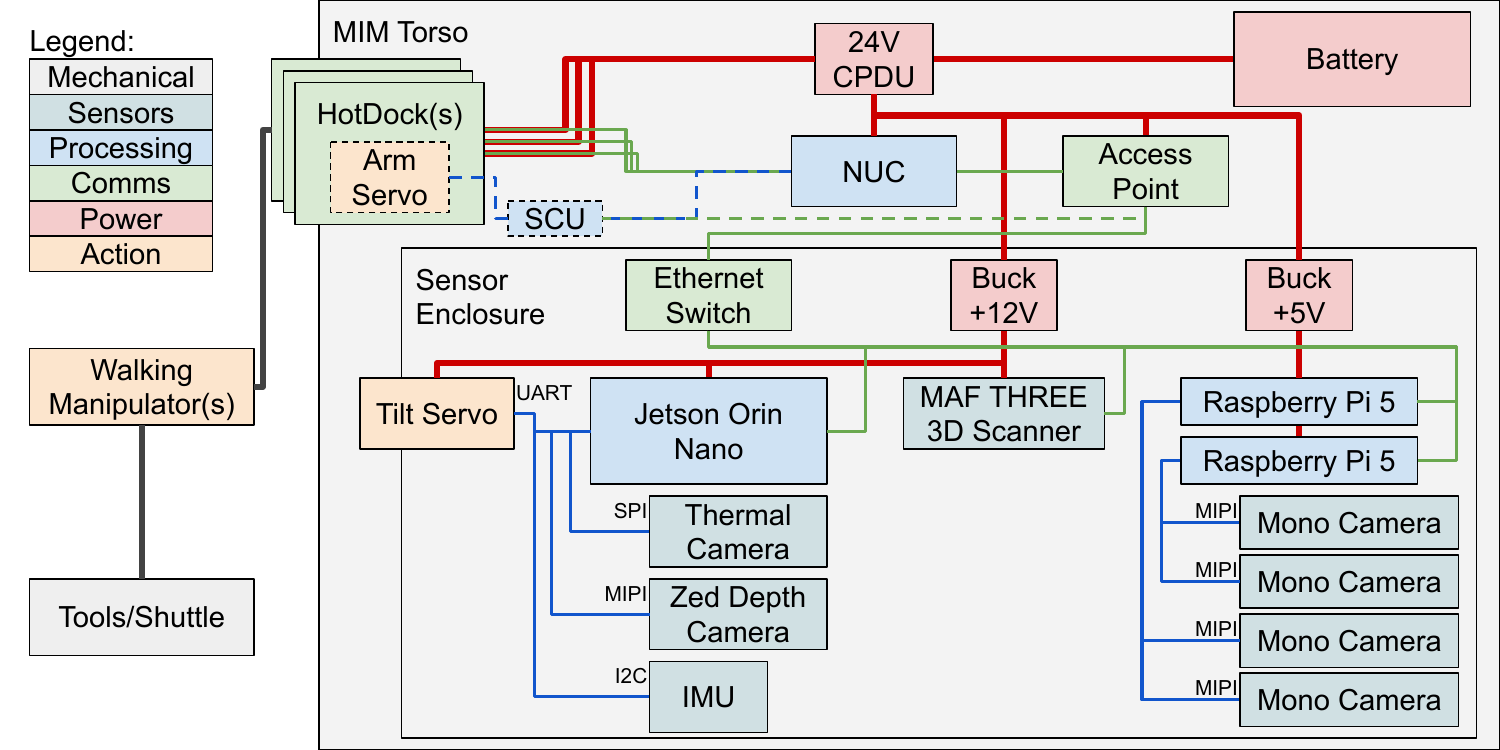}
    \caption{Block diagram of MIM systems with interfacing to external systems.}
    \label{fig:MIM_System_Diagram}
\end{figure}
\vspace{-6pt}
\begin{figure}[tb]
    \centering
    \begin{subfigure}[b]{0.26\textwidth}
        \centering
        \includegraphics[width=\textwidth]{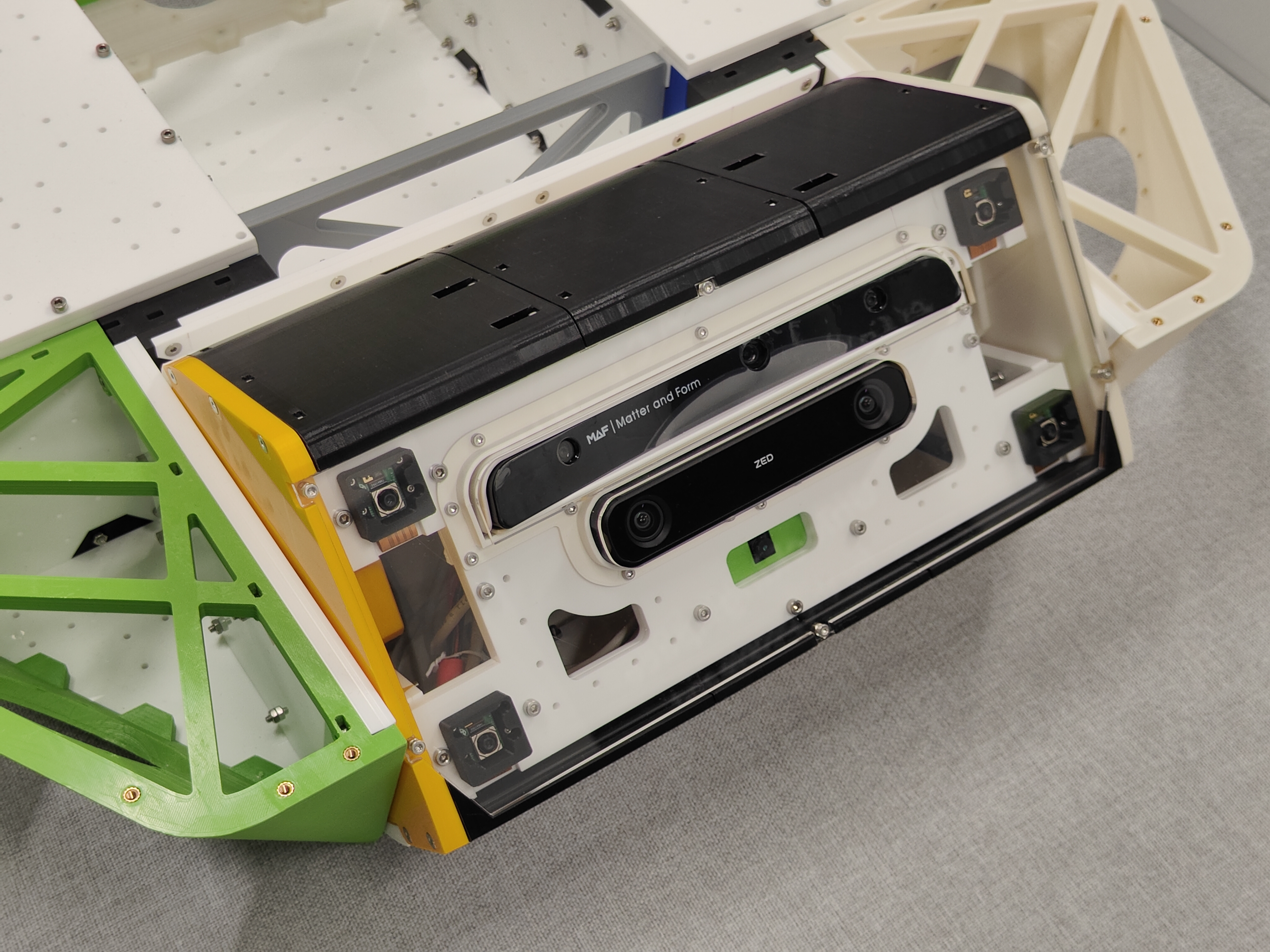}
        \caption{}
        \label{fig:prototype_3d}
    \end{subfigure}
    \begin{subfigure}[b]{0.26\textwidth}
        \centering
        \includegraphics[width=\textwidth]{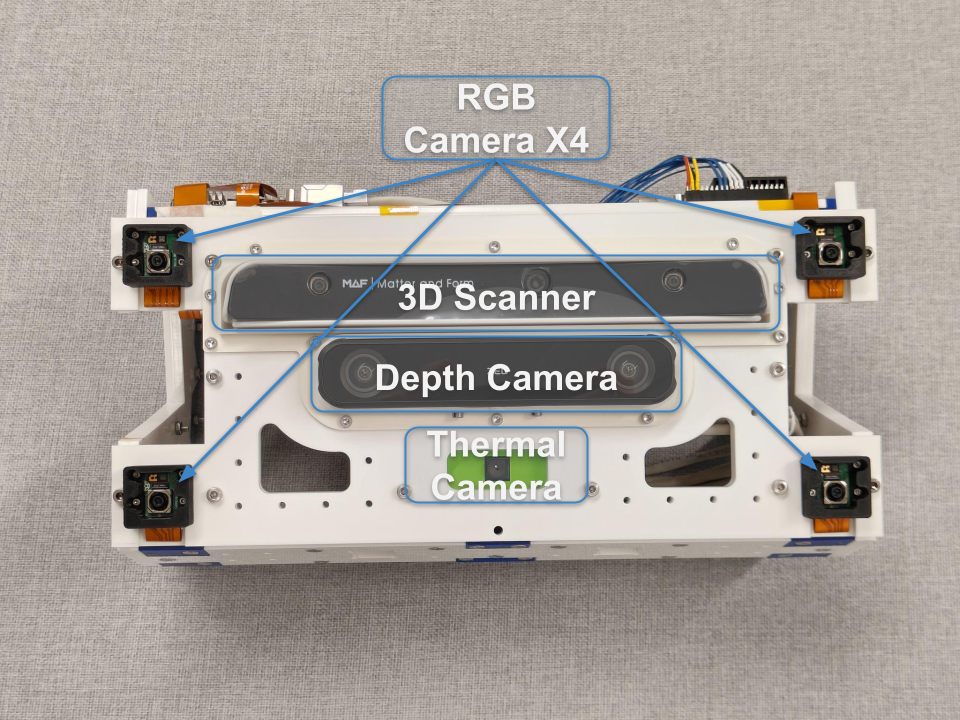}
        \caption{}
        \label{fig:prototype_front}
    \end{subfigure}
    \begin{subfigure}[b]{0.26\textwidth}
        \centering
        \includegraphics[width=\textwidth]{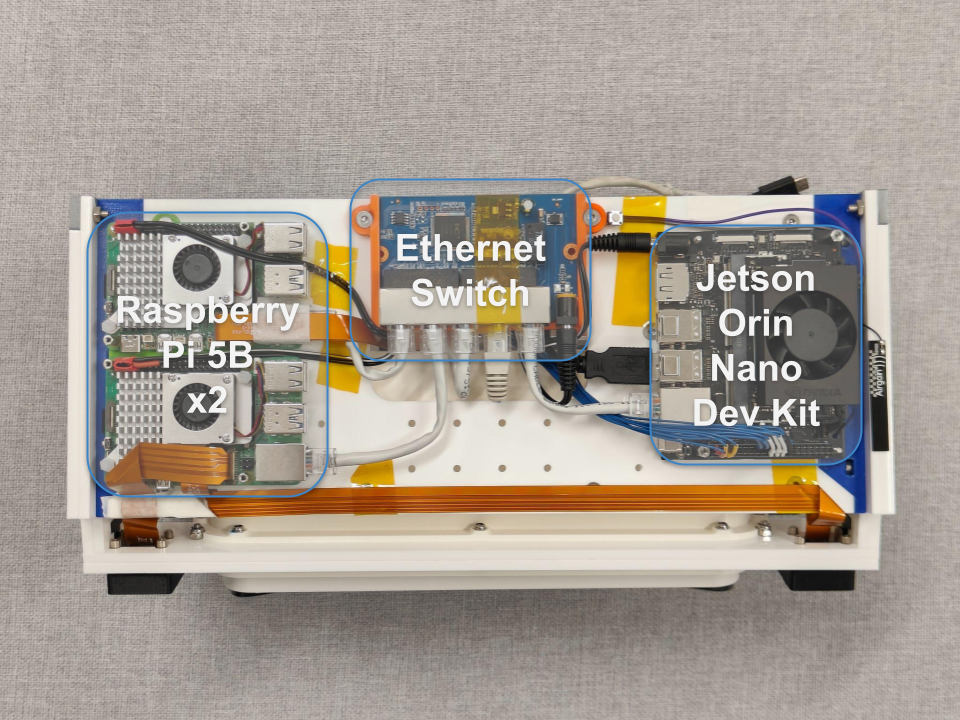}
        \caption{}
        \label{fig:prototype_top}
    \end{subfigure}
    \vspace{-10pt}
    \caption{Prototype of the sensor enclosure and containing perception electronics. From (\textbf{a}) to (\textbf{c}): the 3D, front, and top view of the enclosure and electronics.}
    \label{fig:sensor_prototype}
\end{figure}
\vspace{-6pt}
\begin{figure}[tb]
    \centering
    \begin{subfigure}[b]{0.32\textwidth}
        \centering
        \includegraphics[width=\textwidth]{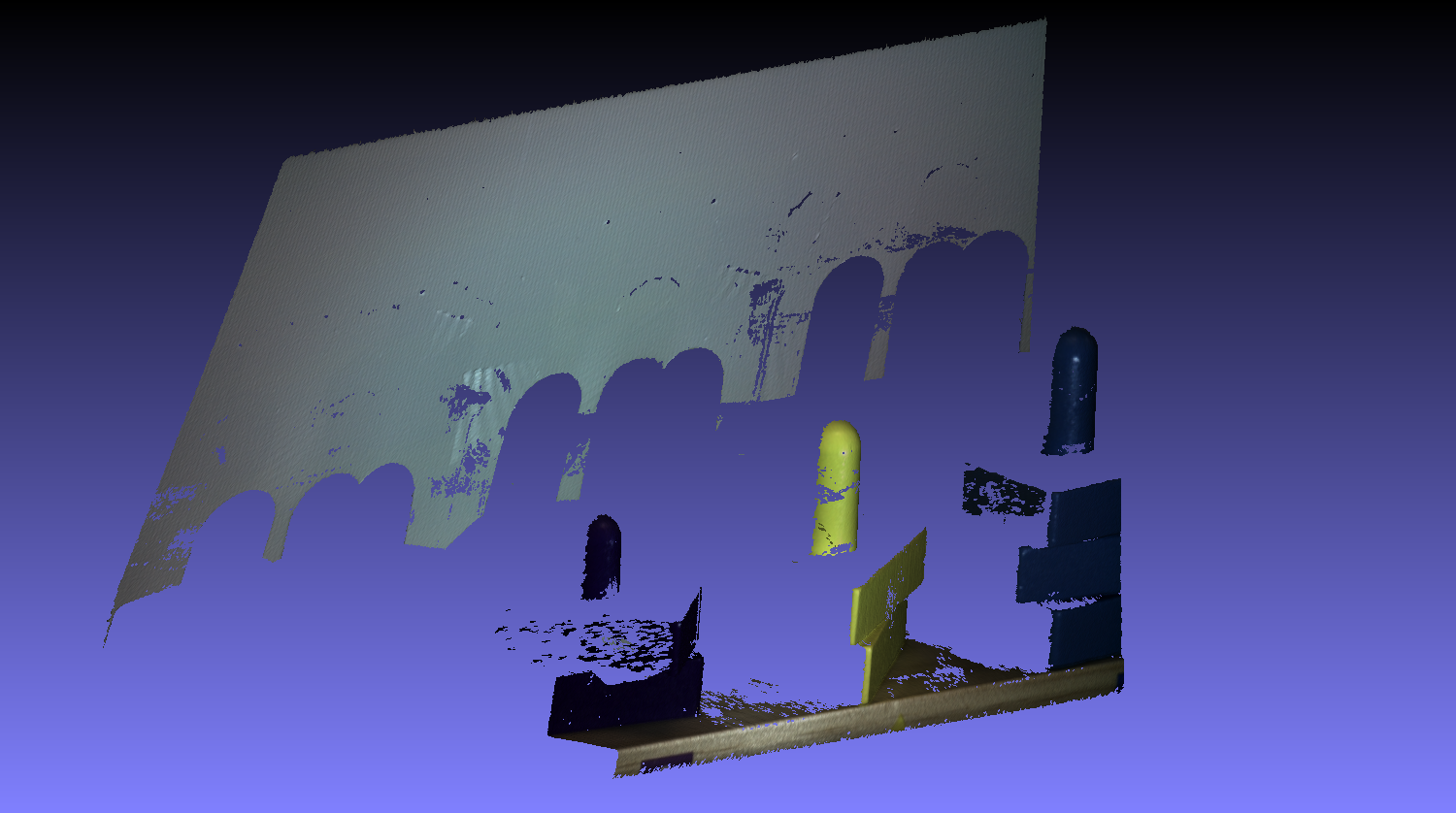}
        \caption{}
        \label{fig:scanner_readout}
    \end{subfigure}
    \begin{subfigure}[b]{0.38\textwidth}
        \centering
        \includegraphics[width=\textwidth]{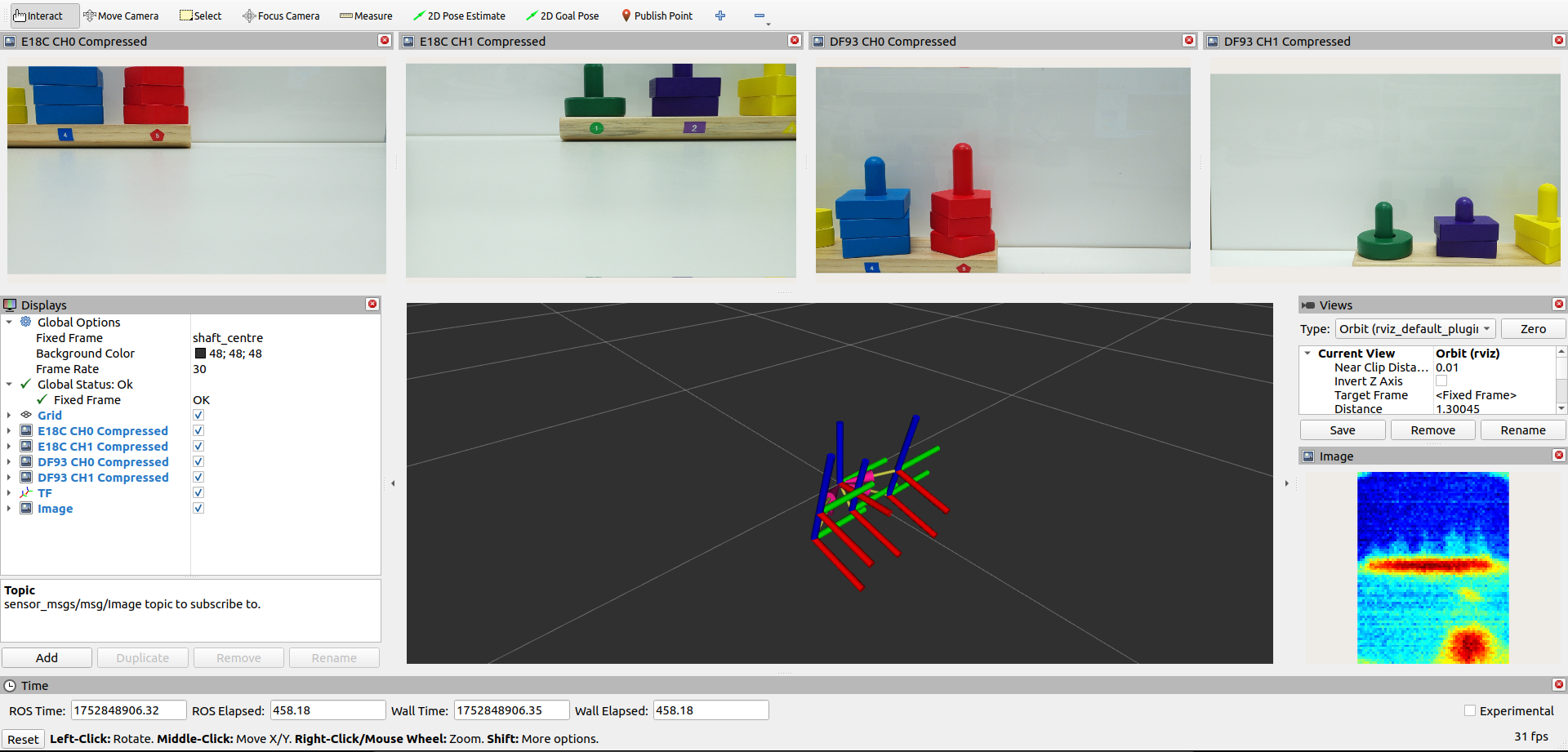}
        \caption{}
        \label{fig:cams_readout}
    \end{subfigure}
    \vspace{-10pt}
    \caption{Initial data from the MIM's onboard sensors. (\textbf{a}) 3D scanner generated mesh model. (\textbf{b}) Sensor readout visualized on the rviz2 interface. }
    \label{fig:sensor_readout}
\end{figure}
\vspace{-4pt}
\subsubsection{Perception Sensors}
To fulfil the I\&M requirements, the sensors in the MIM primarily cover RGB, depth, point cloud, and thermal imaging domains. A block diagram of the system framework and the prototype of the sensor enclosure are shown in Figure \ref{fig:MIM_System_Diagram} and \ref{fig:sensor_prototype}, respectively. 
For RGB images, to ensure the sharpness of objects captured at different distances, camera with a zoom lens and local image processing capability is preferred. For better compatibility with the OBCs and ROS2 communication middleware, the Raspberry Pi 5 with Pi Camera 3 sensors was selected. They are mounted at the four corners of the sensor enclosure's front panel, providing a wide field of view around potential obstructions. Although these cameras are expected to be sufficient for depth imaging, capacity for an accelerated ZED X stereo camera is included for testing purposes.  For sub-millimetre surface profiling, most LIDAR units have insufficient point cloud density, so a Matter And Form THREE 3D structured light camera system is used to achieve sufficient surface resolution while internally handling additional processing overheads.
Coarse resolution thermal imaging is used to detect the presence of operational anomalies that cause lack of activity or overheating. For appropriate dynamic range with a small footprint, a Waveshare Long-wave IR Thermal Imaging Camera with 80×62 resolution is used.  
\vspace{-16pt}
\section{Conclusions}
\vspace{-8pt}
The STARFAB Mobile Inspection Module represents a novel modular approach to inspection and maintenance, suitable for the flexible ecosystems of space hardware components needed to achieve sustainable autonomous operations in Earth orbit and beyond.  The next steps in MIM development are to test the complete prototype in identifying and profiling ORUs and STARFAB structural elements as a platform for algorithm development to achieve the I\&M performance requirements.  Following verification activities, a MIM with aluminium chassis and near-flight capable components will perform autonomous inspection and maintenance as part of the STARFAB project demonstration in mid-2026.
\vspace{-8pt}
\begin{credits}
\subsubsection{\ackname} This research is co-funded by the UK Research and Innovation (UKRI) Horizon Europe guarantee and the European Union through the HORIZON EUROPE Research Programme “Future Space Ecosystem and Enabling Technologies” (Project No. 101135507) entitled “A Space Warehouse Concept and Ecosystem to Energize European ISAM (STARFAB)”.
\vspace{-8pt}
\subsubsection{\discintname}
The authors have no competing interests to declare that are relevant to the content of this article.
\end{credits}
%
%
%
\bibliographystyle{splncs04}
\bibliography{bibliography}

\begin{thebibliography}{1}
\providecommand{\url}[1]{\texttt{#1}}
\providecommand{\urlprefix}{URL }
\providecommand{\doi}[1]{https://doi.org/#1}

\bibitem{deremetz2022design}
Deremetz, M., Debroise, M., De~Stefano, M., Mishra, H., Brunner, B., Grunwald,
  G., Roa~Garzon, M.A., Reiner, M., Z{\'a}vodn{\'\i}k, M., Komarek, M., et~al.:
  Design and integration of a multi-arm installation robot demonstrator for
  orbital large assembly. In: 73rd International Astronautical Congress, IAC
  2022 (2022)

\bibitem{deremetz2024starfab}
Deremetz, M., Letier, P., Gancet, J., Post, M.A., Behling, J., Ullrich, N.,
  Croes, V., Debroise, M., Wouters, M.: Starfab: Concept of operations and
  preliminary system definition for an orbital automated hub and warehouse unit
  supporting in-space operations and services. In: 75th International
  Astronautical Congress (IAC) 2024. International Astronautical Federation
  (IAF) (2024)

\bibitem{dominguez2020common}
Dominguez, R., Post, M., Fabisch, A., Michalec, R., Bissonnette, V.,
  Govindaraj, S.: Common data fusion framework: An open-source common data
  fusion framework for space robotics. International Journal of Advanced
  Robotic Systems  \textbf{17}(2),  1729881420911767 (2020)

\bibitem{letier2024hotdock}
Letier, P., Deremetz, M., Gancet, J., Grunwald, G., Roa, M.A., Brunner, B.,
  Yan, X.T.: Hotdock and the mosar walking manipulator, technologies for
  modular spacecraft assembly and reconfiguration. In: Space Robotics: The
  State of the Art and Future Trends, pp. 515--537. Springer (2024)

\bibitem{horizon-starfab}
\relax Space Applications Services~NV/SA: {STARFAB} a space warehouse concept
  and ecosystem to energize european in-space servicing assembly and
  manufacturing (2024), \url{https://www.horizon-starfab.com/}, accessed:
  2025-02-28

\end{thebibliography}
\end{document}